\pdfoutput=1

\documentclass[11pt]{article}

\usepackage[]{acl}
\usepackage[skins,breakable]{tcolorbox}

\usepackage{times}
\usepackage{latexsym}

\usepackage{graphicx}
\usepackage{subfigure}
\usepackage{amsmath}
\usepackage{multirow}
\usepackage{multicol}
\usepackage{caption}
\usepackage{longtable}
\usepackage{lipsum}

\usepackage[T1]{fontenc}

\usepackage[utf8]{inputenc}

\usepackage{microtype}

\usepackage{hyperref}

\title{SuperRAG: Beyond RAG with Layout-Aware Graph Modeling}

\author{Jeff Yang$^{1}$, Duy-Khanh Vu$^{1}$, \textbf{Minh-Tien Nguyen}$^{2}$\thanks{$^*$Corresponding Author.}, Xuan-Quang Nguyen$^1$, \\\textbf{Linh Nguyen}$^1$, \textbf{Hung Le}$^3$  \\
        $^1$Cinnamon AI, 10th floor, Geleximco building, 36 Hoang Cau, Dong Da, Hanoi, Vietnam. \\
        \texttt{\{jeff.yang, klein, albert, linh\}@cinnamon.is} \\
        $^2$Hung Yen University of Technology and Education, Hung Yen, Vietnam. \\
        \texttt{tiennm@utehy.edu.vn}\\
        $^3$Deakin University, Australia. \\
        \texttt{thai.le@deakin.edu.au}
        }

\begin{document}
\maketitle
\begin{abstract}
This paper introduces layout-aware graph modeling for multimodal RAG.
Different from traditional RAG methods that mostly deal with flat text chunks, the proposed method takes into account the relationship of multimodalities by using a graph structure.
To do that, a graph modeling structure is defined based on document layout parsing.
The structure of an input document is retained with the connection of text chunks, tables, and figures.
This representation allows the method to handle complex questions that require information from multimodalities.
To confirm the efficiency of the graph modeling, a flexible RAG pipeline is developed using robust components.
Experimental results on four benchmark test sets confirm the contribution of the layout-aware modeling for performance improvement of the RAG pipeline.

\end{abstract}

\section{Introduction}
Retrieval Augmented Generation (RAG) \cite{guu2020retrieval,lewis2020retrieval,borgeaud2022improving,izacard2023atlas} is a new paradigm that helps to reduce the hallucination of large language models (LLMs) \cite{cao2020factual,raunak2021curious,ji2023survey} by providing additional contexts for prompting LLMs \cite{su2021prototype,chen2024benchmarking}.
Recently, the approach has gained considerable attention due to its effectiveness in enhancing the capabilities of LLMs \cite{guu2020retrieval,lewis2020retrieval,su2021prototype,xiao2021transductive,borgeaud2022improving,izacard2023atlas}. Within this domain, graph-based RAG has emerged, introducing a novel perspective that leverages structured knowledge to improve further performance and interpretability \cite{panda2024holmes,besta2024multi,li2024graphreader,edge2024local,sun2024docs2kg}.

Unlike non-graph-based RAG methods that directly use raw data as individual chunks of text for downstream reasoning or question-answering tasks, the graph-based RAG approach can represent input data as a graph that considers the relationship among text chunks \cite{panda2024holmes,li2024graphreader,edge2024local}.
We argue that while most RAG-based pipelines perform effectively within the text modality, handling multimodal inputs—common in real-world business applications—poses substantial challenges to these systems, potentially limiting their broader applicability and impact.
The challenge comes from two main reasons.
First, input documents contain diverse layouts, structures, and multimodalities that need to be captured in a RAG pipeline.
The information on the layout plays an important role, helping LLMs understand the document.
Also, the document contains text, tables, and figures which should be encoded into prompts for LLMs' reasoning \cite{zhao2023retrieving}.
Second, an input question may require information in different modalities.
Let's consider the question: ``Please list the standard steps for creating Internet Navigware teaching materials". It requires information in the flow chart on page 27, and text on pages 28, and 29.\footnote{https://software.fujitsu.com/jp/manual/manualfiles/\\m150016/b1ww9681/07z000/tutorial.pdf}



This paper introduces a novel graph-based RAG scheme that addresses the two challenges above for actual multimodal QA cases.
The pipeline includes four steps: document parsing, data modeling, advanced information retrieval, and reasoning.
The document parsing can handle multiple input types using in-house and third-party readers.
For data modeling, we introduce a new knowledge graph (KG) that retains the layout and structure of input documents.
This is because the layout and structure are important to comprehend the meaning of input documents which enhances the performance of the information retrieval (IR) step.
Data modeling in the form of a KG is combined with full-text and vector search to create an advanced IR module that uses re-ranking to retrieve the most relevant contexts.
The combination of multiple retrievers allows the proposed pipeline to retrieve more relevant information from the contexts.
The reasoning step combines an input query and the relevant contexts to form a prompt feed to an LLM for achieving the final answer. 
In summary, this paper makes three main contributions as follows.
\begin{itemize}
    \item It introduces a new Layout-Aware Graph Modeling (LAGM) structure to represent input documents for RAG. The structure is created to retain the layout of input documents which is combined with full-text and vector search to improve the quality of the IR step.

    \item It utilizes state-of-the-art and robust techniques for building a unified RAG pipeline. Experimental results on public benchmark datasets show that the proposed SuperRAG achieves promising results compared to strong other RAG baselines.

    \item It offers a system where users can experience the proposed RAG pipeline (Appendix \ref{sec:app}).
\end{itemize}

\section{Related Work}

\paragraph{RAG}
RAG is a new method that supports LLMs to fill the gap of out-of-date knowledge \cite{he2022rethinking} and hallucination \cite{cao2020factual,raunak2021curious,ji2023survey}.
By using relevant information retrieved from external knowledge, RAG can help LLMs to generate more accurate and reliable responses \cite{guu2020retrieval,lewis2020retrieval,borgeaud2022improving,izacard2023atlas,ren2023investigating,shi2024replug}.
With the aid of RAG, LLMs have achieved promising results in many tasks such as code generation \cite{zhou2022docprompting}, domain-specific QA \cite{cui2023chatlaw,dahl2024large,pu2024autorepo}, or open-domain QA \cite{izacard2021leveraging,trivedi2023interleaving,kim2024sure,wang2024knowledge,yu2024visrag}.


\paragraph{Graph-based RAG}
The graph structure has been adapted to capture relationships among concepts such as Connected Papers tool,\footnote{https://www.connectedpapers.com} a tree of summary nodes for long context \cite{chen2023walking}, or multimodal KGs for storing text, diagrams, and source code \cite{kannan2020multimodal}.
The graph has also been used to improve the quality of RAG in different ways such as hyper-relational KG \cite{panda2024holmes}, graph-based agents for long contexts \cite{li2024graphreader}, KG for summarization \cite{edge2024local}, or graph neural networks \cite{mavromatis2024gnn}.
However, we observed that most of these efforts have focused on the text modelity.



We follow the direction of building multimodal KGs for RAG \cite{sun2024docs2kg,wang2024knowledge}.
While prior works have explored hierarchical document parsing for RAG, SuperRAG differentiates by emphasizing structured granularity and document layout analysis.
We introduce a modern, generalized data model, incorporating Table of Contents (ToC) and master sections to improve retrieval for large documents.
These enhancements preserve document structure, enhancing retrieval accuracy and effectiveness.
We also share the idea of using the structure of documents for RAG \cite{saad2023pdftriage}; however, our method empowered by an in-house reader that can handle diverse document types with table and chart understanding rather than only processing the text structure of PDF files as \citet{saad2023pdftriage}.

\section{Layout-Aware Graph Modeling}


Layout-aware Graph Modeling (LAGM) is designed to effectively represent input documents while preserving their original layout and structure. This approach is motivated by the need to enhance the comprehensibility and manageability of property graphs, particularly for applications involving multimodal and complex data.
For example, if the query asks for information in a table or chart, the RAG pipeline needs to know which section or subsection it belongs to.
\subsection{Document Layout Parsing}
The first step in constructing LAGM is parsing input documents using specialized readers for different modalities, including text, tables, diagrams, and images. This step outputs a structured format that forms the foundation for graph creation. We leverage an in-house document parser with the enhancement from Azure DI to ensure robust processing across diverse layouts.

\begin{figure}[!h]
\centering
\includegraphics[scale=0.14]{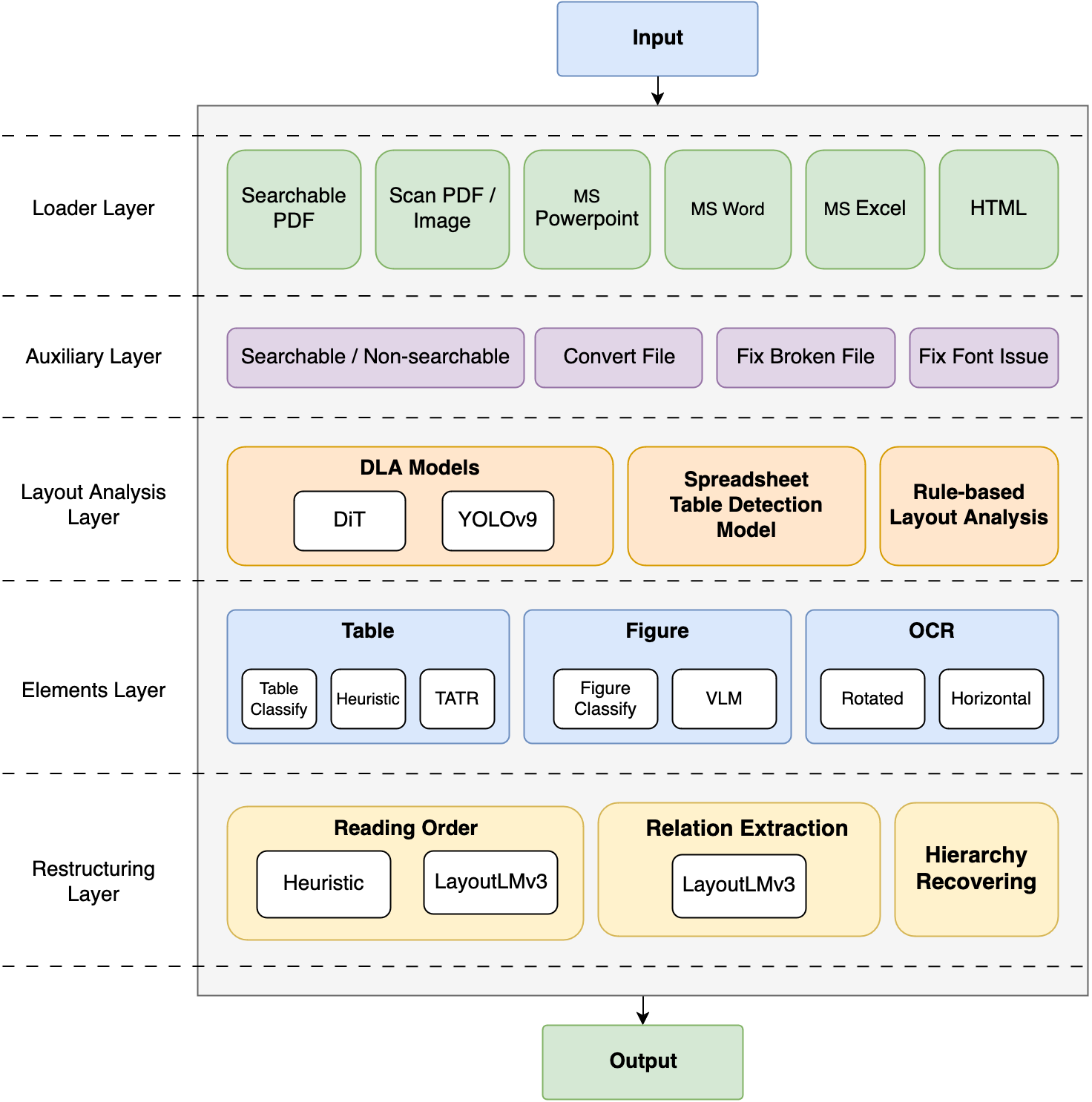}

\caption{The pipeline of the in-house parser.}\label{fig:parser}
\end{figure}

\paragraph{The In-House Document Parser}
Our in-house parser is designed as a modular pipeline to process each page independently (Figure \ref{fig:parser}). It begins with a loader layer for format conversion and pre-processing, followed by AI models for extracting layouts, table structures, OCR, and figure content. The processed data undergoes post-processing, such as reading order sorting and relation extraction, and is output in JSON/Markdown.

Key components of the in-house parser include Document Layout Analysis (DLA), reading order detection, table structure recognition, and figure and table classification. The DLA module is pre-trained on DocLayNet \cite{pfitzmann2206doclaynet} and further fine-tuned with 5773 in-house annotated PDF pages, enabling the model to recognize 9 distinct layout labels like titles, tables, and figures. 

For reading order detection, the parser employs the method proposed by \citet{wang2021layoutreader}, leveraging 5010 annotated document images to extract natural reading sequences. Table structure recognition is implemented using an in-house library designed to identify diverse table formats accurately. Lastly, figure and table classification rely on a curated dataset to categorize tables into subtypes (e.g., full-lined, borderless) and figures into specific types (e.g., charts, diagrams), ensuring precise extraction of visual elements.
\begin{table}[!h]
\centering
\small
\caption{Document reading performance.}
\begin{tabular}{lccc}
\hline
\textbf{Methods} & \textbf{NID} & \textbf{TEDS} & \textbf{TEDS-S} \\ \hline

Amazon Textract & 96.71 & 88.05 & 90.79 \\ 
LlamaParse & \textbf{92.82} & 74.57 & 76.34 \\ 
Unstructured & 91.18 & 65.56 & 70.00 \\ 
Google Layout Parser & 90.86 & 66.13 & 71.58 \\ 
Azure DI & 87.69 & 87.19 & 89.75 \\ 
Our reader DI & 92.43 & \textbf{89.76} & \textbf{91.14} \\ \hline

\end{tabular}
\label{tab:reader-compared}
\end{table}
Table \ref{tab:reader-compared} reports the comparison of the in-house reader with orther strong reading methods.
\textbf{NID} stands for Normalized Indel Distance for layout and order
reading.
\textbf{TEDS} is Tree Edit Distance-based Similarity for
text and table structure recognition.
TEDS-S is Tree Edit Distance-based Similarity-Struct for table structure recognition only.
We can observe that the in-house reader achieves competitive results which are good to implement actual RAG pipelines.

\paragraph{Azure DI for PDF Parsing Enhancement}
Azure DI enhances the parser by excelling in section-header and paragraph detection. It supports searchable and non-searchable PDFs and aids in creating ToC. 
To generate the ToC, we use Azure DI outputs for tables, sections, and diagrams, performing the following:  (1) Match physical and printed page numbers.  (2) Detect ToC based on keywords.  (3) Replace printed page numbers with physical page numbers.  This integration ensures superior layout-aware graph modeling and improves ToC generation for structured navigation.

\subsection{Data modeling}
\begin{figure}[!h]
\centering
\includegraphics[scale=0.5]{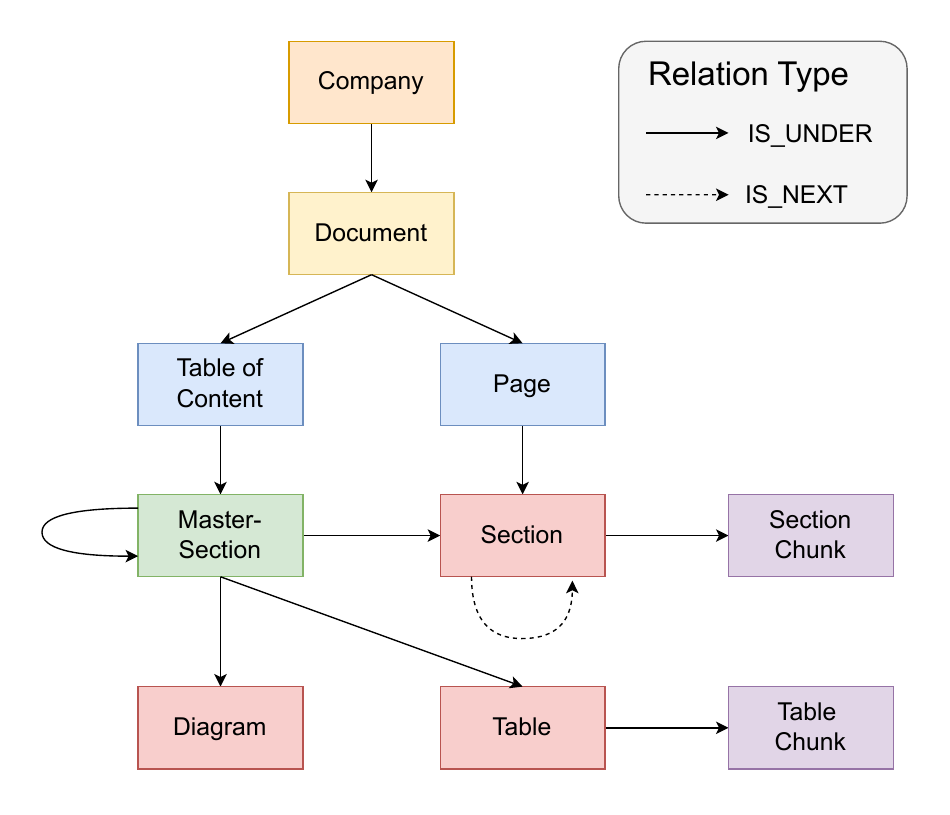}

\caption{The knowledge graph used for data modeling.}\label{fig:kg}
\end{figure}

After parsing, each document page can be decomposed into title, header, sections, text chunks, tables and diagrams, etc.
The data modeling step aims to create a granular-level design for the property graph.
Figure \ref{fig:kg} shows the definition of LAGM.

The \textbf{Company} node serves as the root, representing the overarching entity or corpus, such as a company, and capturing metadata like the company's name. Each \textbf{Document} node, linked to the Company, represents an individual document with attributes such as document name, type, and path.

Documents connect to \textbf{Page} nodes, which represent individual pages and include attributes like page index, headers, footers, and textual content. The \textbf{TableOfContents} node, also linked to Document, provides a structural overview of the document and connects to \textbf{MasterSection} nodes. MasterSections organize the content hierarchically and link to \textbf{Section}, \textbf{Table}, and \textbf{Diagram} nodes.

\textbf{Section} nodes represent logical divisions within a document and include attributes like section headers and content. Sections are connected sequentially via ``\texttt{has\_next}" relationships, ensuring the flow of content. They can also link to finer-grained \textbf{SectionChunk} nodes, capturing texts under the section. \textbf{Table} nodes, representing tabular data, and \textbf{Diagram} nodes, representing visual elements, provide additional structure. Tables may be further connected \textbf{TableChunk} nodes for storing textual contents inside the table.
These explicit ``\texttt{is\_under}" and ``\texttt{has\_next}" relationships reflect the natural hierarchy and flow of documents. This design supports layout-aware graph modeling and efficient information retrieval, enhancing applications like RAG pipelines by enabling precise navigation and knowledge extraction.

\subsection{The SuperRAG Framework}

Building on layout-aware graph modeling (LAGM), we introduce an advanced retrieval expansion framework combining LLM-based and heuristic-driven approaches for flexible and efficient information retrieval. This framework enhances RAG-based pipelines by improving adaptability and scalability across applications.

\paragraph{LLM-Based Graph Traversal.}  
This approach leverages a Large Language Model (LLM) to perform context-aware graph traversal. Using the graph schema (visualized in Fig. \ref{fig:kg}) as input, the LLM dynamically generates Cypher queries, enabling intelligent and relationship-driven retrieval. It is particularly effective for complex, multimodal data and intricate document structures encoded in the graph. Detailed information of the prompt for the LLM is mentioned at the end of the appendix.

\paragraph{Heuristic-Based Retrieval.}  
Complementing the LLM-based approach, the framework processes ToC, tables, and diagrams as heuristics for IR enhancement.
For ToC, the framework uses structured output from the LLM with prompt engineering (Fig. \ref{fig:toc}) and heuristics to extract the ToC during indexing.
This is because ToC contains important structured information for retrieval.
During retrieval, it computes semantic similarity scores between section titles and the query for targeted content retrieval.
Additionally, few-shot prompting is used to ask the LLM to directly extract the relevant page based on a given query.
For table processing, the DETR model \cite{carion2020end} for table detection and recognition is used, followed by an OCR engine to reconstruct the table structure before indexing.
This ensures that tables are accurately captured and searchable within the SuperRAG pipeline.
For diagram processing, OCR models are used to extract text from diagrams and feed both images and text information into a multimodal LLM (e.g., GPT-4o) for better interpretation.
This allows context-aware understanding of visual elements, ensuring better integration of diagrams in retrieval and reasoning.
These methods are computationally efficient, effective, and robust for dealing with structured content.


\paragraph{Comparative Insights.}  
The dual framework balances flexibility and efficiency, with LLM-based traversal excelling in unstructured, exploratory tasks, and heuristics providing predictable performance for high-throughput systems. Together, they enable scalable and adaptive RAG pipelines, leveraging graph structures for optimal retrieval.

\begin{figure*}[!h]
\centering
\includegraphics[width=0.8\textwidth]{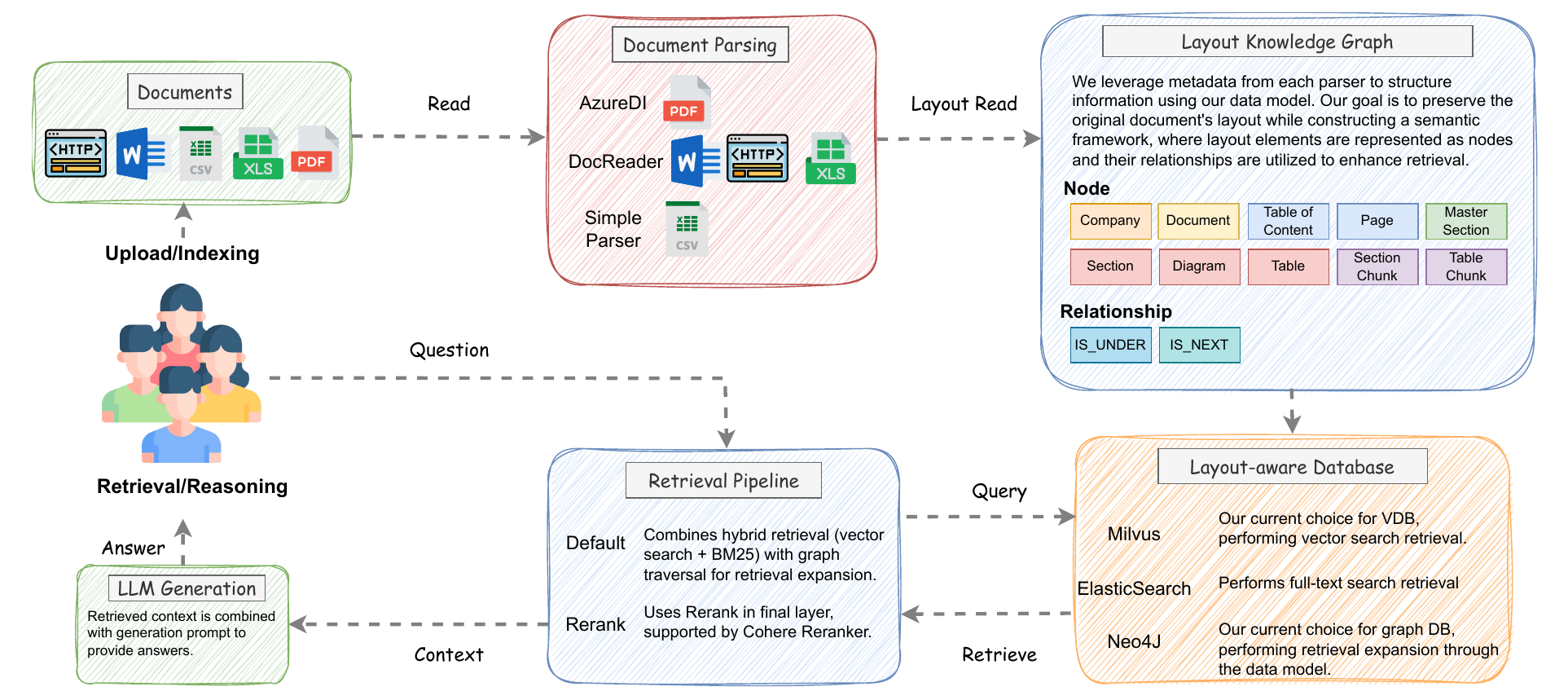}
\caption{The proposed SuperRAG framework.}\label{fig:gr}
\end{figure*}

\subsection{Graph Augmentation}

To enrich the LAGM, we employ the $K$-Nearest Neighbors (KNN) \cite{cover1967nearest} as a graph augmentation technique to create new 
\texttt{is\_similar} relationships between nodes within the graph. The KNN algorithm calculates similarity between nodes based on their properties, using metrics such as cosine similarity, Jaccard similarity, or Euclidean distance, depending on the data type.
Also, \texttt{has\_stem} relationships are generated using synonyms or words sharing the same stem, linking nodes representing conceptually related terms.

\section{Applications}
Figure \ref{fig:gr} shows the pipeline of LAGM that integrates multiple retrievers and re-rankers, combining heuristic graph traversal, similarity search, and language model-based techniques for efficient retrieval and ranking.
The pipeline is flexible in several ways. First, it merges cross-page context using the graph representation. Second, a TOC retriever is included for documents with structured information, improving context quality for specific queries. Additionally, the pipeline uses diagram/table expansion for queries needing information from tables and diagrams, with a self-reflection layer to evaluate whether table or diagram information is necessary based on the query intent.
It selectively integrates these elements only when they contribute to a more accurate answer, reducing irrelevant content retrieval.
Notably, LAGM is pipeline-agnostic and can integrate into any RAG pipeline.

\section{Experimental Settings}
\subsection{Datasets}
We examine the following datasets for evaluation.

\paragraph{DOCBENCH} is a benchmark designed to evaluate LLM-based document reading systems \cite{zou2024docbenchbenchmarkevaluatingllmbased}. It features 1,102 questions and 229 PDF documents from five domains: academia, finance, government, laws, and news, with an average of 66 pages and 46,377 tokens per document.

\paragraph{SPIQA} includes 27K research papers in three tasks: direct QA with figures and tables, direct QA with full papers, and CoT QA. The evaluation contains test-A (666 filtered questions), test-B (228 human-written questions from QASA), and test-C (493 from QASPER), all emphasizing reasoning with figures and tables.

\begin{table*}[!h]
\centering
\caption{The comparison on DOCBENCH.}
\begin{tabular}{lcccccccccc}
\hline
\textbf{System} & \textbf{Aca} & \textbf{Fin} & \textbf{Gov} & \textbf{Laws} & \textbf{News} & \textbf{Text} & \textbf{Multi} & \textbf{Meta} & \textbf{Una} & \textbf{Avg. Acc} \\ \hline
\multicolumn{11}{c}{Layout-aware vs. non-layout-aware data modeling} \\ \hline
Non-layout & 64.0 & 70.1 & 64.2 & 62.8 & 83.7 & 77.7 & 74.4 & 46.1 & 70.2 & 68.5 \\ 
Layout-aware & \textbf{75.9} & \textbf{79.9} & \textbf{71.6} & \textbf{65.4} & 83.7 & \textbf{84.7} & \textbf{85.1} & \textbf{50.4} & \textbf{75.8} & \textbf{75.8} \\ \hline

\multicolumn{11}{c}{Layout-aware vs. SOTA RAG methods} \\ \hline
GPT4 (API) & 65.7 & 65.3 & 75.7 & 69.6 & 79.6 & \textbf{87.9} & 74.7 & 50.8 & 37.1 & 69.8 \\ 
GPT-4o (API) & 56.4 & 56.3 & 73.0 & 65.5 & 75.0 & 85.0 & 62.7 & 50.4 & 17.7 & 63.1 \\ 
KimiChat (Web) & 62.4 & 61.8 & \textbf{77.0} & 78.5 & 87.2 & 87.6 & 65.3 & 50.4 & 71.8 & 70.9 \\ 
Claude 3 Opus (Web) & 73.9 & 40.6 & 70.3 & \textbf{79.1} & \textbf{86.6} & 80.8 & 64.6 & \textbf{54.3} & 58.9 & 67.6 \\ 
\hline
SuperRAG (Ours) & \textbf{75.9} & \textbf{79.9} & 71.6 & 65.4 & 83.7 & 84.7 & \textbf{85.1} & 50.4 & \textbf{75.8} & \textbf{75.8} \\ \hline
\end{tabular}\vspace{-0.2cm}
\label{tab:results}
\end{table*}

\subsection{Detailed Implementation}
Milvus was used as a vector database.
ElasticSearch was used for full-text search.
Neo4J was implemented as a graph database.
The embedding model uses embedding-v3-large from Open AI.
LLM for completion uses GPT-4o with version 2024-05-01.
The hyper-parameters include selecting the top 3 tables and diagrams, the top 20 for relevant contexts, and the top 10 for re-ranking.

\subsection{Evaluation Metrics}
All models were assessed using a GPT-4-based evaluator, which has demonstrated a 98\% agreement with human annotators, ensuring robust and reliable accuracy measurement \cite{zou2024docbenchbenchmarkevaluatingllmbased}.

\section{Results and Discussion}
This section first reports the performance comparison of SuperRAG with other strong RAG-based methods, and then shows the ablation study, output observation.
It finally describes the demo system.

\subsection{Performance on RAG Tasks}
\paragraph{Layout-aware vs. non-layout-aware}
The first comparison includes two settings: layout-aware and non-layout-aware.
The layout-aware approach leverages document structure—such as headers, tables, figures, and sections—to provide contextual cues that are often critical for accurately understanding and retrieving information across varied domains. In contrast, the non-layout-aware model only uses Hybrid Search for IR with a flat structure.

The first part of Table \ref{tab:results}, and Table \ref{tab:results2}, demonstrate that layout-aware modeling significantly enhances performance across domains and tasks. On DOCBENCH, the layout-aware model achieves an average accuracy of 75.8\%, outperforming the non-layout model’s 68.5\% by 7.3 points. Notably, in academia and finance, gains are 11.9 and 9.8 points, respectively, showing the value of structural cues in complex documents.
On SPIQA in Table \ref{tab:results2}, the layout-aware model improves Test-A accuracy by 4.5 points (59.\% vs. 55.4\%) and Test-B by 1.3 points (63.1\% vs. 61.8\%). In the challenging Test-C, it achieves an average accuracy gain of 9 points (57.2\% vs. 48.2\%), with notable improvement in table handling. These results confirm layout awareness as a key factor in improving contextual understanding and retrieval accuracy.

\begin{table}[!h]
\centering
\caption{Layout-aware vs. non-layout-aware on SPIQA Test-B and Test-C. ColPali is used for Qwen 2B, 7B, Claude, and GPT-o4.}
\begin{tabular}{lccc}
\hline
\textbf{System} & \textbf{Figure} & \textbf{Table} & \textbf{Avg. Acc} \\ \hline
\multicolumn{4}{c}{Test-A} \\ \hline
Non-layout & 53.9 & 57.2 & 55.4 \\ 
Layout-aware & \textbf{57.4} & \textbf{63.7} & \textbf{59.9} \\ \hline

\multicolumn{4}{c}{Test-B} \\ \hline
Non-layout & 62.4 & 61.0 & 61.8 \\ 
Layout-aware & \textbf{66.1} & \textbf{58.9} & \textbf{63.1} \\ \hline

\multicolumn{4}{c}{Test-C} \\ \hline
Non-layout & 57.5 & 44.6 & 48.2 \\  
Layout-aware & \textbf{58.2} & \textbf{56.7} & \textbf{57.2} \\ \hline

\end{tabular}
\label{tab:results2}
\end{table}

\paragraph{Comparison with SOTA methods}
The proposed data modeling was compared to state-of-the-art RAG methods.
On DOCBENCH, we compare our method against state-of-the-art LLM-based document reading systems, including proprietary pipelines like GPT-4, KimiChat, and Claude-3. For SPIQA, since the benchmarked results only measure baseline QA performance using full gold context without including the IR component of the RAG system, a direct comparison would be unfair. To address this, we reran several strong baselines using a full IR pipeline instead of relying on reported numbers from original papers. Additionally, we evaluated ColPali \cite{faysse2024colpali}, an open-source retrieval model that generates contextualized embeddings from document page images, contrasting with our layout-focused method.

As shown in the second part of Table \ref{tab:results} and Table \ref{tab:results4}, our approach SuperRAG consistently outperforms other systems across both DOCBENCH and SPIQA benchmarks. On DOCBENCH, SuperRAG achieves the highest overall accuracy (75.8\%), particularly excelling in the Financial and multi-type questions. In comparison, proprietary systems like GPT-4 and KimiChat perform strongly in specific categories, but their overall accuracies fall short by at least 6\% compared to our method. Notably, SuperRAG's ability to handle a wide range of question types, especially complex multi-type and una-type questions, highlights its superior document comprehension capabilities.
\begin{table}[!h]
\centering
\caption{The performance on SPIQA Test-B and Test-C. ColPali is used for Qwen 2B, 7B, Claude-3.5 Sonnet.}
\begin{tabular}{lccc}
\hline
\textbf{System} & \textbf{Figure} & \textbf{Table} & \textbf{Avg. Acc} \\ \hline
\multicolumn{4}{c}{Test-A} \\ \hline
GPT-4o (API) & 51.6 & 54.2 & 52.7 \\ 
Qwen 2-7B & 48.3 & 40.5 & 45.9  \\ 
Claude-3.5 Sonnet & \textbf{58.1} & 56.8 & 57.6 \\
SuperRAG (Ours) & 57.4 & \textbf{63.5}  & \textbf{59.9} \\ \hline

\multicolumn{4}{c}{Test-B} \\ \hline
GPT-4o (API) & 63.1 & 53.6 & 59.2 \\ 
Qwen 2-7B & 41.3 & 45.2 & 42.9  \\ 
Claude-3.5 Sonnet & 53.3 & 44.2 & 49.5 \\ 
SuperRAG (Ours) & \textbf{66.2} & \textbf{58.9} & \textbf{63.2} \\ \hline

\multicolumn{4}{c}{Test-C} \\ \hline
GPT-4o (API) & 43.1 & 40.9 & 41.5 \\ 
Qwen 2-7B & 40.2 & 28.5 & 31.8  \\ 
Claude-3.5 Sonnet & 46.0 & 42.3 & 43.4 \\
SuperRAG (Ours) & \textbf{58.2} & \textbf{56.7} & \textbf{57.2} \\ \hline

\end{tabular}
\label{tab:results4}
\end{table}

For SPIQA, SuperRAG demonstrates superior performance across all three test sets, excelling in both figure and table-based QA tasks. In Test-A, it achieves the highest average accuracy (59.9\%), with a notable 63.5\% on table-based questions, outperforming the best baseline by 7\%. For Test-B, SuperRAG again leads with an average accuracy of 63.2\%, surpassing the strongest baseline Claude3.5 Sonet (49.5\%). It achieves 66.2\% on figure-related tasks and 58.9\% on table-based tasks, showcasing balanced strengths across modalities. In Test-C, SuperRAG achieves 57.2\% overall, with standout performances in both figures (58.2\%) and tables (56.7\%). In comparison, the runner-up Claude-3.5 Sonnet trails at 46.0\%, marking a substantial gap of 12.2\%. These results underscore SuperRAG’s ability to handle multimodal inputs effectively, even when competing with enterprise systems.


\subsection{Ablation Study}
We investigate the flexibility of the pipeline by testing with three settings.
The first setting is the non-layout method which uses the hybrid search + cross-page context merger (\textbf{1}).
The second setting is the layout-aware method which uses the hybrid search + cross-page context merger + TOC integration + table-diagram expansion (\textbf{2}).
The TOC integration is to extract the Table-of-Content in documents.
The table-diagram expansion expands the context with tables and diagrams relevant to the input query.
The final setting is also our proposed layout-aware method which is similar to the second setting but using self-reflection (\textbf{3}).
Self-reflection means that the pipeline decides whether to use information from tables and diagram expansion based on the input query.

Table \ref{tab:ablation} presents the accuracy results across various settings. Our method, equipped with all functionalities, consistently achieves the highest accuracy, highlighting the effectiveness of each component in enhancing overall system performance.

\begin{table}[ht]
\centering
\caption{Component contribution. DOC: DOCBENCH.}
\begin{tabular}{ccccc}
\hline
\textbf{Setting} & \textbf{DOC} & \textbf{Test-A} & \textbf{Test-B} & \textbf{Test-C} \\ \hline

1 & 68.5 & 55.4 & 61.8 & 48.2 \\ 
2 & 71.7 & 53.0 & 60.9 & 53.1 \\
3 & \textbf{75.8} & \textbf{59.9} & \textbf{63.1} & \textbf{57.2} \\ \hline

\end{tabular}\vspace{-0.4cm}
\label{tab:ablation}
\end{table}

\subsection{Output Observation}
The performance of RAG pipelines was observed to show their behavior on raw samples.
To do that, the observation was done with three methods: non-layout, layout-aware, and ColPali (using Sonnet).
Tables \ref{tab:output1} and \ref{tab:output2} show the outputs of the three pipelines.
For the first sample in Table \ref{tab:output1}, the non-layout-aware pipeline could not output correct answer.
This is because it could not retrieve correct relevant context for RAG.
The ColPali method gives an uncertain answer because the rank of the paper retriever page image from Colpali (topk=1 or topk=3) does not contain enough information and the reasoning capability on the image of VLM still have some disadvantage.
The layout-aware gives the correct answer (retrieval information from both images (in page 2 - Reference 2 in Page 3) and text content from page 3 and another page).
It shows the efficiency of the proposed layout-aware method for retrieving relevant context.
For the second sample in Table \ref{tab:output2}, both layout and non-layout model are all based on the benchmark tables for accuracy data and cannot retrieve information about test errors in figure \texttt{d}.
The ColPali method can not retrieve extract page contain figure \texttt{d} with top 1 or top 3.
As the result, it could not output a correct answer.
In this case, all the RAG pipelines could not retrieve the figure \texttt{d}.
I suggests that the retrieval of visual components in documents should be improved.

\subsection{The Demo System}
Figure \ref{fig:system} provides an interface where users can experience the system.
The right panel includes settings for uploading files, IR types, and other settings.
The central panel consists of a text box for inputting queries.
After putting a query, the system retrieves relevant context based on the layout-aware graph modeling and responses the final answer.
The right panel provides evidence of the answer, that contains confidence scores and relevant chunks.
Related information is highlighted in the relevant chunks.
The open source version can be found at \texttt{\small https://github.com/Cinnamon/kotaemon}.


\section{Conclusion}
The paper introduces layout-aware graph modeling for multimodal data construction used by RAG.
The modeling takes into account the structure of input documents for building a graph that contains the relationship among text chunks, tables, and figures.
A RAG pipeline has also been developed to confirm the efficiency of the modeling.
Experimental results on four public test sets show two important points.
First, layout-aware modeling is beneficial for improving the performance of RAG compared to non-layout-aware and strong other RAG pipelines.
Second, the designed RAG pipeline is flexible, and adding more sophisticated RAG-related components improves the performance of the system.
The modeling and RAG pipeline are practical for business scenarios.


\section*{Limitations}
First, our approach relies heavily on accurate document layout parsing and high-quality data modeling. If these components are misaligned or if document structure extraction tools are limited, the pipeline’s effectiveness may be reduced. In particular, noisy layouts or variations in document structures across domains could impact the quality of information retrieval (IR) and subsequently the reasoning performance of the pipeline. Moreover, integrating tables, figures, and non-text elements into a coherent graph structure may introduce computational overhead, making the pipeline resource-intensive. This can affect scalability, especially in real-world applications requiring high throughput or settings with limited computational resources.

\section*{Ethics Statement}
Our framework presents no major ethical concerns, as it has been designed with a genuine focus on improving the accuracy of information retrieval in LLM-based systems. Our method does not generate or alter content independently but instead organizes multimodal information from existing documents, ensuring that outputs remain faithful to the source material. Privacy risks are minimized by following data protection regulations and implementing strict anonymization protocols where needed, particularly for sensitive data.

\bibliographystyle{acl_natbib}
\bibliography{custom}

\section*{Appendix}\label{sec:app}


\paragraph{Prompt for LLM-based Graph Traversal}

\begin{figure}[!h]
\centering
\includegraphics[width=0.5\textwidth]{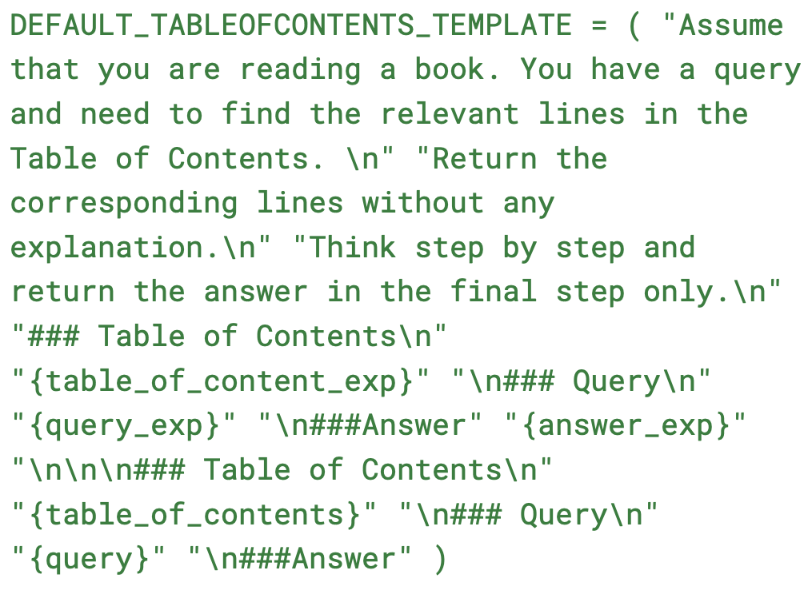}
\caption{The proposed SuperRAG framework.}\label{fig:toc}
\end{figure}

The ToC prompt example is shown in Fig. \ref{fig:toc}.

An example prompt for LLMs to generate Cypher graph queries is included at the end of the appendix (\ref{box:cypher}).

\paragraph{The output observation}
The examples of output observation are shown in Tables \ref{tab:output1} and \ref{tab:output2}.
\begin{table*}[ht]
\centering
\small
\caption{Output observation on two samples of SPIQA.}\label{tab:output}
\begin{tabular}{p{15cm}}
\hline
\textbf{Question:}  Which of the two approaches, density estimation or mixture of classifiers, is more robust to outliers? (https://arxiv.org/pdf/1805.06431v4) \\
\textbf{Correct answer:} Mixture of classifiers \\ 
\textbf{Evidence:} The figure shows that the mixture of classifiers approach is able to correctly classify the data points even when there are outliers present. This is because the mixture of classifiers approach is able to learn the different modes of the data distribution, while the density estimation approach is not. \\ \hline
\textbf{Non-layout:} Based on the provided context, particularly from References 4, 6, and 7, the proposed method (ChoiceNet) incorporates a mixture of densities to robustly learn the underlying target distribution in the presence of noisy (corrupt) training data, whereas the mixture of classifiers is also discussed but seems to face challenges in distinguishing clean and corrupt patterns independently.

In the regression tasks and synthetic examples described in References 4 and 7, it is shown that the proposed method, which is based on a mixture density network, outperforms or shows comparable results with low outlier rates and demonstrates superior performance with high outlier rates (greater than 50\%). This is evident from the comparison with other methods, including a standard mixture density network (MDN) and other robust loss methods. Specifically, Figures 6(a)-(c) in Reference 8 illustrate that the proposed method maintains lower average fitting errors across different outlier rates, and Figures 6(d)-(f) in Reference 8 show better fitting results even with 60\% outliers.

On the other hand, the mixture of classifiers approach, while theoretically sound, faces practical challenges. As stated in Reference 3, the mixture of classifiers might not work well in practice because both classifiers can independently learn corrupt patterns, leading to a mixture weight adhering to 1/2 and making it hard to distinguish clean from corrupt data without additional dependencies or regularization.

Therefore, based on the empirical evidence and detailed explanations given in the provided context, the density estimation approach (specifically the proposed mixture density network method) is more robust to outliers compared to the mixture of classifiers approach.  \\ \hline
\textbf{Claude 3 Opus:} Based on the information provided in the image, it's difficult to definitively state which approach - density estimation or mixture of classifiers - is more robust to outliers. The figure shows both approaches being used as part of a process for binary classification on corrupt data, but doesn't explicitly compare their robustness to outliers.

However, we can make some inferences:

1. The mixture of classifiers approach (part b of the figure) seems to separate the data points more clearly, with distinct decision boundaries (shown as red lines labeled W). This could potentially make it more robust in handling outliers or corrupt data points.

2. The density estimation approach (part a) shows overlapping distributions of data points. While this may capture the overall distribution well, it might be more susceptible to being influenced by outliers.

3. The text mentions that the mixture of classifiers approach (using $\hat{W}$) is "expected to reveal the reverse patterns by minimizing the following mixture classification loss". This suggests that this approach might be designed to better handle corrupt data or outliers.

4. The motivation section discusses how the mixture approach aims to "discriminate the corrupt data by using another linear classifier $\hat{W}$", which could imply better handling of outliers.

Given these points, there's a slight indication that the mixture of classifiers approach might be more robust to outliers. However, to make a definitive statement, we would need more specific information about how each method handles outliers, which isn't explicitly provided in the image or text.
  \\ \hline
\textbf{Layout-aware:} The approach described in the document for handling outliers involves two primary methods: density estimation and mixture of classifiers. Based on the provided references, the robustness to outliers can be inferred from the empirical results and the theoretical framework described.

1. **Density Estimation Approach**:
   - The density estimation approach leverages a mixture density network (MDN) to model the target conditional density using a parameterized distribution, typically a Gaussian distribution (Reference 3).
   - The robustness of this approach is evaluated in regression tasks with synthetic data and real-world datasets like the Boston Housing dataset (Reference 4). The results show that while the proposed method, built on top of an MDN, outperforms or shows comparable results with low outlier rates, it exhibits superior performance with high outlier rates (> 50\%) (Reference 4).
   - Additionally, the MDN's performance degrades severely with extreme noise levels (60\%), indicating that without dependencies among different mixture components, the MDN struggles to allocate high mixture probability correctly (Reference 4).

2. **Mixture of Classifiers Approach**:
   - The mixture of classifiers approach involves modeling the choice probability of different classifiers and estimating the corruption probability using the mixture weight \( T_w(x) \) (Reference 2).
   - This method employs a linear functional with negative correlation between classifiers to help distinguish clean and corrupt patterns (Reference 3).
   - An illustrative example in a binary classification task using the MNIST dataset shows that ChoiceNet (which uses a mixture of classifiers) successfully infers the clean data distribution and effectively handles corrupted labels (Reference 7).
   - The final test accuracies reported (81.7\% train accuracy and 98.1\% test accuracy) suggest that this approach can robustly handle noisy labels (Reference 7).

**Comparison and Conclusion**:
- The density estimation approach shows that it can handle high outlier rates, but its performance degrades significantly with extreme noise levels.
- The mixture of classifiers approach, particularly as implemented in ChoiceNet, demonstrates strong performance in distinguishing clean from corrupted data and maintains high accuracy even with a significant proportion of noisy labels.

Based on the empirical results and the robustness demonstrated, **the mixture of classifiers approach appears to be more robust to outliers** compared to the density estimation approach. The mixture of classifiers method effectively identifies and manages corrupted data through its correlation-guided mechanism, leading to better handling of noisy datasets.  \\ \hline

\end{tabular}
\label{tab:output1}
\end{table*}

\begin{table*}[ht]
\centering
\small
\caption{Output observation of the second sample from SPIQA.}
\begin{tabular}{p{15cm}}
\hline

\textbf{Question:}  Which model performs the best in terms of test error? (https://arxiv.org/pdf/1805.06431v4) \\
\textbf{Correct answer:} CDAN (M) \\ 
\textbf{Evidence:} Figure (d) shows the test error for different models. CDAN (M) has the lowest test error. \\ \hline
\textbf{Non-layout:} Based on the provided context, the model that performs the best in terms of test error is the "CDAN+E (w/o random sampling)" model. This model achieves the highest average accuracy across all transfer tasks, as shown in the table in Reference 8. Specifically, it achieves an average accuracy of 87.7\%, which is higher than the other variants of CDAN+E (with Gaussian sampling and with uniform sampling).

The performance across specific tasks such as A -> W, D -> W, W -> D, A -> D, D -> A, and W -> A also supports this conclusion, where the CDAN+E (w/o random sampling) consistently shows superior or comparable performance in comparison to the other models. Therefore, it can be concluded that "CDAN+E (w/o random sampling)" performs the best in terms of test error among the models listed.
  \\ 
\textbf{Claude 3 Opus:} I apologize, but I don't see any information in this image about model performance or test errors. The image contains technical details about generalization error analysis, domain adaptation theory, and descriptions of various datasets used in experiments. It does not show any results or comparisons of model performance. To answer which model performs best in terms of test error, we would need to see actual experimental results or performance metrics, which are not provided in this particular image.  \\ \hline

\textbf{Layout-aware:} Based on the provided context, particularly from Reference 1 (page 7), the model that performs the best in terms of test error across various datasets and tasks is the **CDAN+E** model. This can be observed from the consistently high classification accuracies reported for CDAN+E across multiple benchmark datasets when compared to other models.

Here is a detailed breakdown from the tables:

1. **Office-31 Dataset (AlexNet and ResNet-50 backbones):**
   - CDAN+E achieves the highest average accuracies of 77.7\% (AlexNet) and 87.7\% (ResNet-50).

2. **ImageCLEF-DA Dataset:**
   - CDAN+E records an average accuracy of 80.3\% (AlexNet) and 87.7\% (ResNet-50), again being the top performer.

3. **Office-Home Dataset:**
   - CDAN+E shows the highest average accuracy of 65.8\% (ResNet-50).

4. **Digits Dataset (MNIST, USPS, SVHN):**
   - CDAN+E achieves an average accuracy of 94.3\%, outperforming other methods.

5. **VisDA-2017 Synthetic to Real Dataset:**
   - CDAN+E achieves an accuracy of 70.0\%, which is higher than other comparison methods.

Therefore, across these datasets, CDAN+E consistently outperforms other domain adaptation methods, indicating it has the lowest test error and is the best-performing model overall. \\ \hline

\end{tabular}
\label{tab:output2}
\end{table*}

\begin{figure*}[!h]
\centering
\includegraphics[scale=0.35]{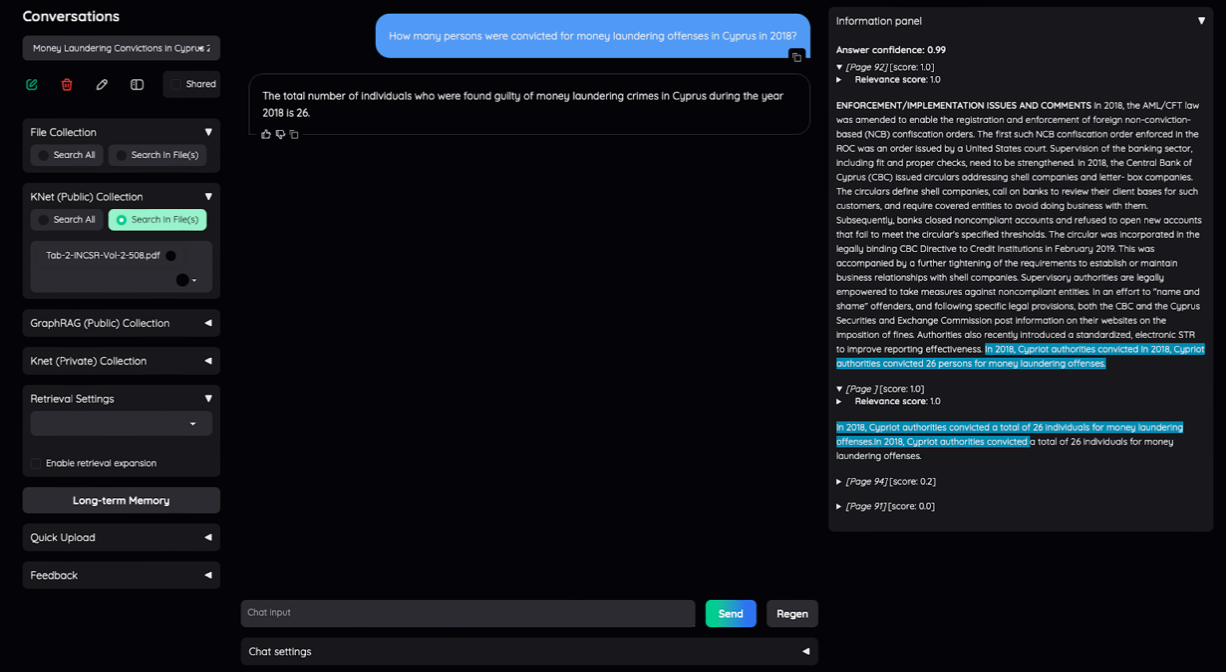}

\caption{The demo system with the sample from the DOCBENCH dataset. The input question is \textit{``How many persons were convicted for money laundering offenses in Cyprus in 2018?"} and the answer is \textit{``26 persons"}.}\label{fig:system}
\end{figure*}

\paragraph{The demo system}
The user interface of the system is shown in Fig. \ref{fig:system}.

\begin{tcolorbox}[
    enhanced, 
    breakable,
    float*,width=\textwidth,
     colback=gray!5, colframe=blue!50!black, 
     before=\par\smallskip\noindent, after=\par\smallskip]
     \small
            \texttt{
CYPHER\_QUERY\_TEMPLATES = """You are required to construct a Cypher query to retrieve the requested information from the graph database. The graph schema is provided below for reference.\\
\{graph\_schema\}\\
Instructions for Cypher Query Generation:\\
1. Schema Adherence:\\
\quad - Use only the provided relationship types and properties.\\
2. Response Guidelines:\\
\quad - Generate a Cypher query as plain text without any additional formatting.\\
\quad - Include only the Cypher statement; exclude any explanations, apologies, or unrelated content.\\
3. Conditions for Query Construction:\\
\quad - Use \texttt{pageIdx} and \texttt{parentPageIdx} to identify the page. Do not use \texttt{pageNumber}.\\
\quad - Use the \texttt{docType} attribute to identify the document type.\\
\quad - If \texttt{docName} is provided, use it to filter nodes.\\
4. Handling Uncertainty:\\
\quad - If unsure about the user's request or if no Cypher query is applicable, return nothing.\\
5. Things to Avoid:\\
\quad - Do not generate generic queries. If the request lacks specifics, return nothing.\\
\quad - Do not use or infer any additional relationship types or properties.\\
\quad - Don't generate overly complex queries. Keep the queries simple and focused on the user's request.\\
\quad - Don't generate keyword queries unless explicitly requested.\\
\quad - Don't write queries that could return all \texttt{SECTION}, \texttt{TABLE}, or \texttt{DIAGRAM} nodes from the document.\\
Good Examples:\\
\texttt{-----}\\
\texttt{MATCH (s)-[:S\_IS\_UNDER\_P]->(p:PAGE)}\\
\texttt{WHERE toString(p.pageIdx) IN \$pages AND s.parentDocName IN \$doc\_id}\\
\texttt{RETURN s;}\\
 \texttt{-----}\\
Bad Examples:\\
\texttt{-----}\\
\texttt{MATCH (s:SECTION)}\\
\texttt{WHERE s.parentDocName IN ['<dir>', '<doc\_name>']}\\
\texttt{RETURN s;}\\
\texttt{-----}\\
\texttt{MATCH (s:SECTION)-[:S\_IS\_UNDER\_P]->(p:PAGE)}\\
\texttt{WHERE s.parentDocName IN ['<dir>', '<doc\_name>']}\\
\texttt{RETURN s;}\\
\texttt{-----}\\
User Request: \{user\_request\}\\
docName: \{doc\_name\}\\
Cypher Query (Generate a Cypher query as plain text without any additional formatting):"""\\
            }
        \label{box:cypher}
\end{tcolorbox}

\end{document}